\documentclass[twocolumn,notitlepage,10pt,aip,jcp]{revtex4-1}
\usepackage{setspace}
\usepackage{amsfonts, amsmath, amssymb, verbatim, setspace, pdfsync, graphicx,comment,epsfig}
\usepackage[utf8]{inputenc}

\usepackage{dcolumn}    
\usepackage{bm}         
\usepackage{hyperref}
\usepackage{color}

\usepackage{hyperref}
\usepackage{epstopdf}
\hypersetup{colorlinks,%
citecolor=black,%
filecolor=black,%
linkcolor=black,%
urlcolor=black,%
pdftex}

\begin{document}

\title{Spatio-Temporal Activation Function To Map Complex Dynamical Systems} 
\author{Parth~Mahendra} 
\noaffiliation
\noaffiliation
\affiliation{\mbox{Reading School, Erleigh Road, Reading, Berkshire, RG1 5LW, UK.}}
\begin{abstract}
Most of the real world is governed by complex and chaotic dynamical systems. All of these dynamical systems pose a challenge in modelling them using neural networks. Currently, reservoir computing, which is a subset of recurrent neural networks, is actively used to simulate complex dynamical systems. In this work, a two dimensional activation function is proposed which includes an additional temporal term to impart dynamic behaviour on its output. The inclusion of a temporal term alters the fundamental nature of an activation function, it provides capability to capture the complex dynamics of time series data without relying on recurrent neural networks.
\footnote{From 27th September 2020, Department of Computer Science, University of Warwick, CV4 7AL, UK.}
\end{abstract}
\maketitle
\vspace{-20mm}
\section{Introduction}
Most of the real world is governed by complex and chaotic dynamical systems, vibration engineering, cardiac arrhythmia and stock prices. All of these dynamical systems pose a challenge in modelling via using neural networks.\\
There are several types of neural networks, such as feed-forward, CNN (Convolutional Neural Networks), RNN (Recurrent Neural Networks), Autoencoders, etc. Feed-forward and convolutional neural networks are made up of neurons connected in layers, which take inputs from the previous layer and pass on the output to the next layer. Recurrent neural networks differ from both feed-forward and convolutional neural networks as the neurons in them are connected in a “recurrent” manner, and may have feedback loops between the sets of neurons. The feedback loop allows the network additional capability to capture temporal dynamics and learn time series data. Reservoir computing, which is a subset of recurrent neural networks, was first proposed by Jaeger \cite{jaeger}. In the reservoir computing approach\cite{jaeg-haas,herzog}, the large pool of neurons act as a “reservoir” where the input data is imparted a dynamic behaviour, to generate output. \\
The primary motivation behind this research is to model complex dynamics without relying on a recurrent neural network by including a temporal part in an activation function. The activation function fluctuates based on a time-dependent parameter that can be generated in different ways such as the logistic or cubic map.
\begin{figure}
\begin{center}
{\includegraphics[height=12.5cm]{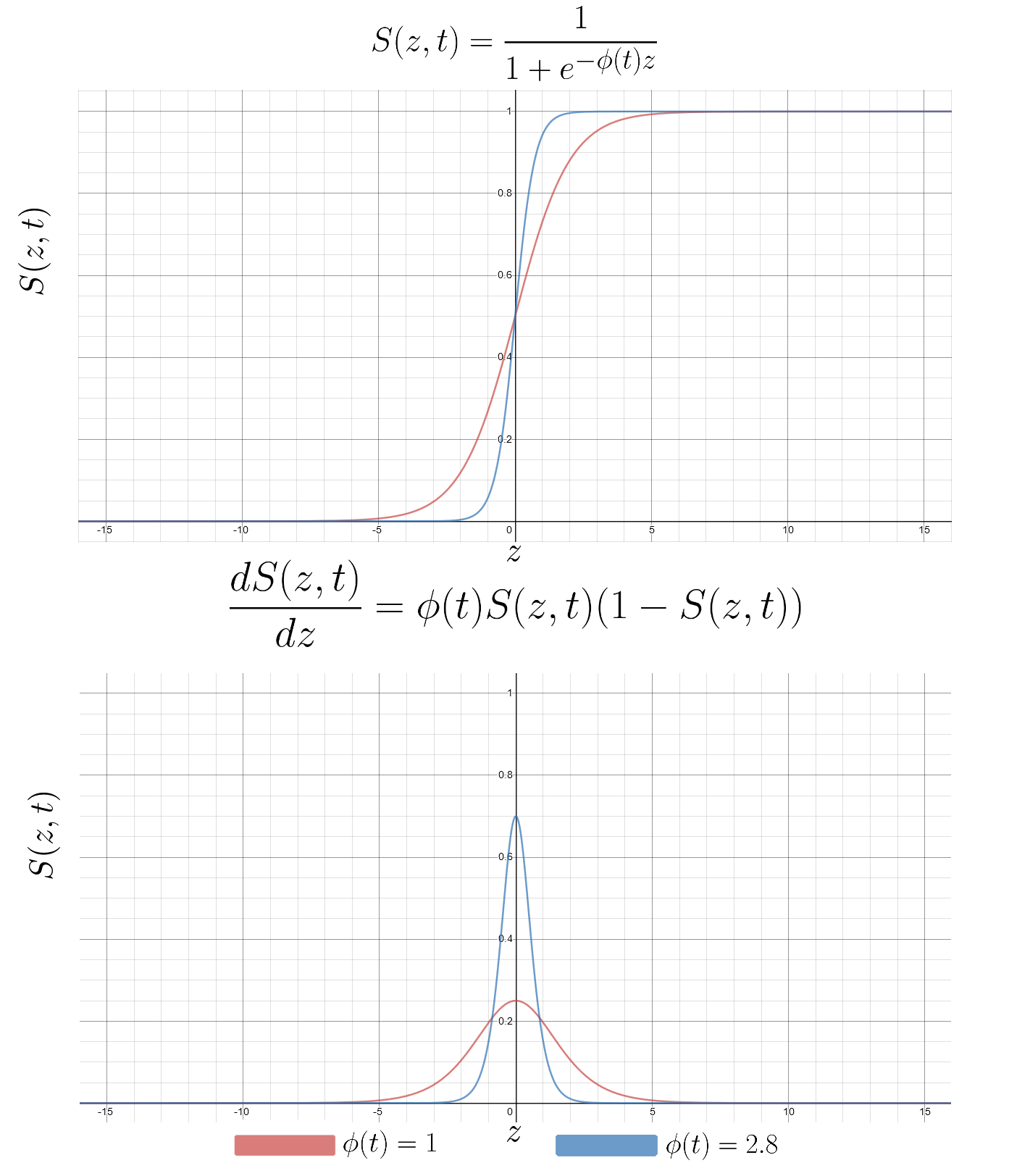}}
\end{center}
\caption{Activation Function and its derivative for $\phi(t)=1$, $\phi(t)=2.8$.}\label{fszd}
\end{figure}
\section{Spatio-Temporal Activation Function}\label{lgnn}
\begin{figure}
\begin{center}
{\includegraphics[height=3cm]{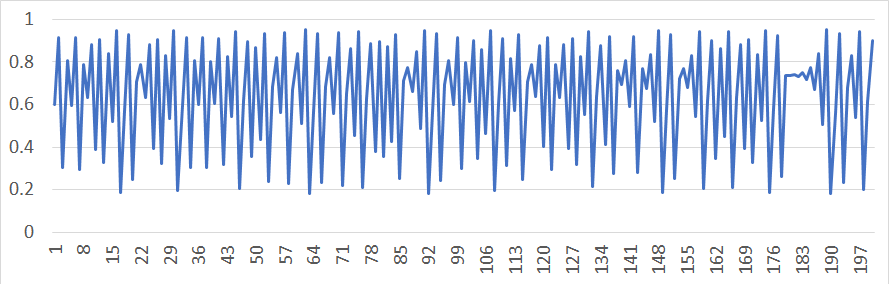}}
\end{center}
\caption{Chaotic dynamical plot of $\alpha(t)$ for $r=4 $.}\label{flog}
\end{figure}

\begin{figure}
	\begin{center}
		{\includegraphics[height=10cm]{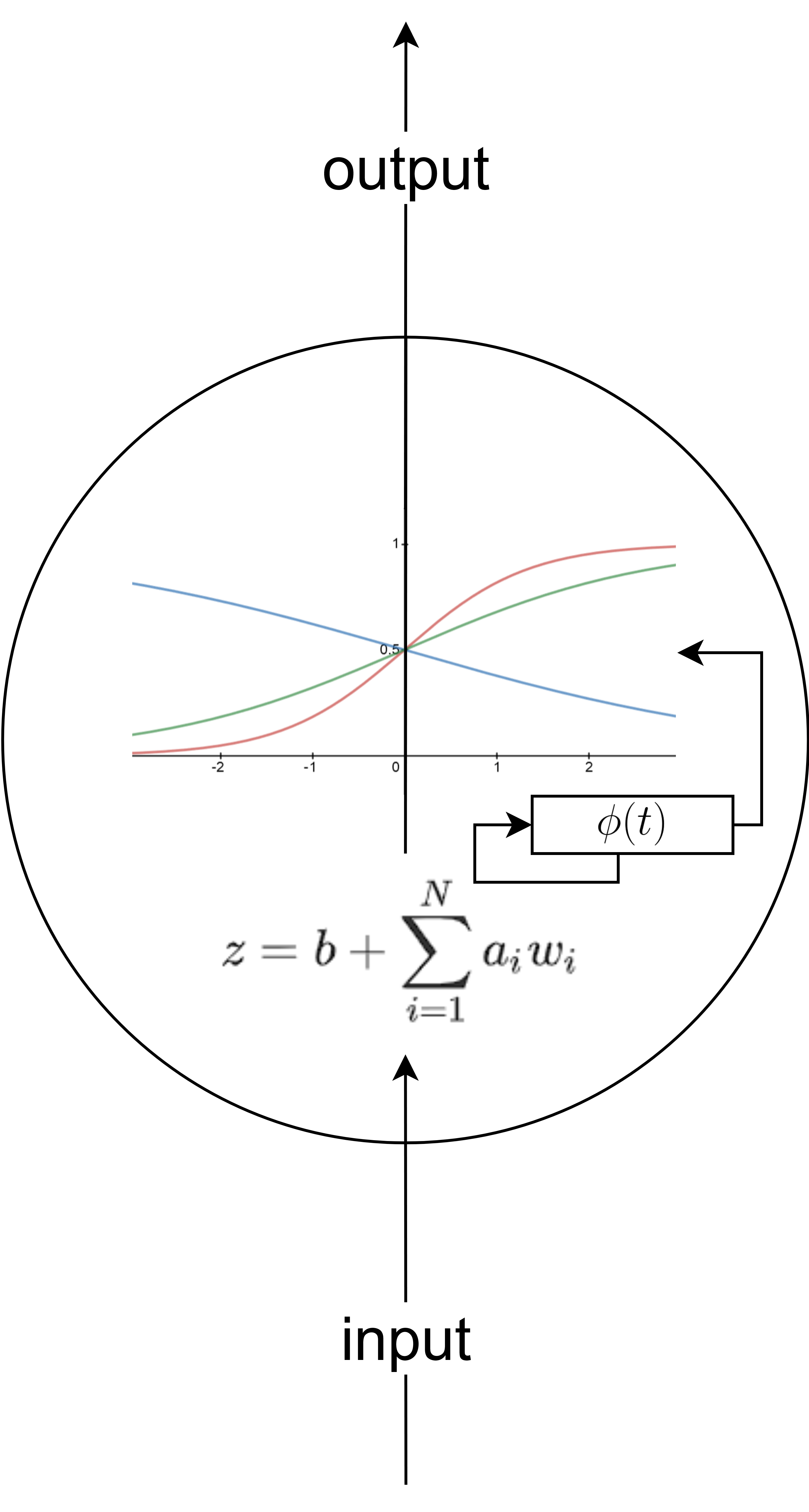}}
	\end{center}
	\caption{Neuron with the spatio-temporal activation function}\label{diagram}
\end{figure}

Each type of network may differ in structural topology, but all share the same basic building block - the neuron and an activation function. In neural networks, the activation function acts as a way to map the weighted sum of the inputs going into each neuron to a useful output. There are four major types of activation functions: binary step, sigmoid, tanh, and ReLU. The neural networks vary in different ways, for example, the sigmoid function generates output that is normalised between $0$ and $1$, while tanh is normalised between $-1$ and $1$. 
In the this research work, a new spatio-temporal activation function, similar to the sigmoid activation function, is proposed, which is a function of the sum of weighted inputs, $z$, and also of a temporal term. The proposed spatio-temporal function $S(z,t)$ is 
\begin{equation} \label{esz} 
S(z,t) = \frac{1}{1-exp(-\phi(t)z)}
\end{equation}
The derivative of  $S(z,t)$ being
\begin{equation} \label{eszd}
\frac{dS(z, t)}{dz}=\phi(t)S(z)(1-S(z))
\end{equation}
The derivative of the new activation function, $S(z,t)$, is similar to the Gaussian probability distribution, and the temporal term $\phi(t)$ plays the role of varying its standard deviation. Figure \ref{fszd} shows the activation function and its derivative for constant $\phi(t)=1$ and $\phi(t)=2.8$. The temporal term $\phi(t)$ in the present case is a linear function of a time-dependant parameter, $\alpha(t)$, as follows
 \begin{equation} \label{newequation}
 \phi(t) = \phi_0+k\alpha(t)
 \end{equation} 
 where $\phi_0$ and $k$ are constants. Figure \ref{diagram} shows the schematic of the activation function.\\
  In this case, $\alpha(t)$ can follow the dynamics of the logistic map \cite{salis} based on the growth parameter $r$ as follows
\begin{equation}\label{elog1}
\alpha (t+1) = r \alpha(t)(1-\alpha(t))
\end{equation}
The function $f(\alpha)=r\alpha(1-\alpha)$ is the logistic map, $f:[0,1] \mapsto[0,1]$, where $3.6<r<4$, giving rise to chaotic dynamical behaviour as shown in Figure \ref{flog}.
It should be noted that Eqn. \ref{elog1} is the time derivative of $\alpha(t)$ while Eqn. \ref{eszd} is the derivative of the sigmoid function, and both follow a similar functional form, one in time domain $t$ and other in input domain $z$. Interestingly, the derivative of the activation function in Eqn. \ref{eszd} is also the	 logistic map.
\begin{figure}
	\begin{center}
		{\includegraphics[height=6cm]{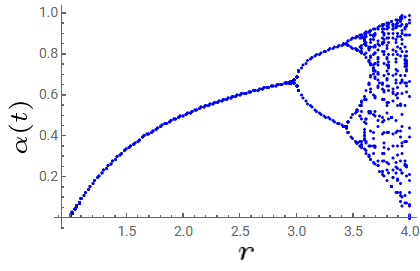}}
	\end{center}
	\caption{Bifurcation diagram of the logistic map}\label{bifurcation}
\end{figure}
\\The temporal function $\phi(t)$ is normalised between $(\phi_{min},\phi_{max})$ as follows
\begin{equation}\label{enorm}
\phi(t) = \phi_{min} + \frac{(\alpha(t) -\alpha_{min})}{(\alpha_{max} - \alpha_{min})}(\phi_{max}-\phi_{min})
\end{equation}
where $\alpha_{max}$ and $\alpha_{min}$ are the maximum and minimum value taken by function $\alpha(t)$.
\section{Response of spatio-temporal activation function}
\begin{figure}
	\begin{center}
		{\includegraphics[height=4.9cm]{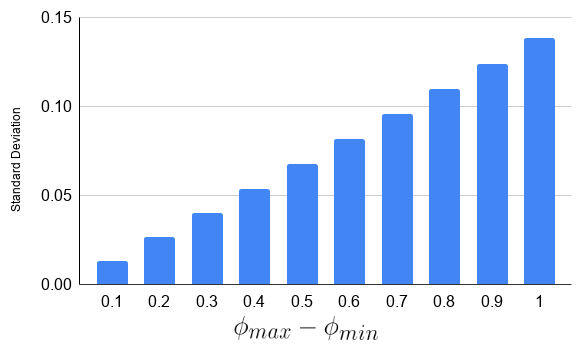}}
	\end{center}
	\caption{Standard deviation corresponding to different ranges of  $(\phi_{max} - \phi_{min})$.}\label{standard}
\end{figure}

\begin{figure}
	\begin{center}
		{\includegraphics[height=4.5cm]{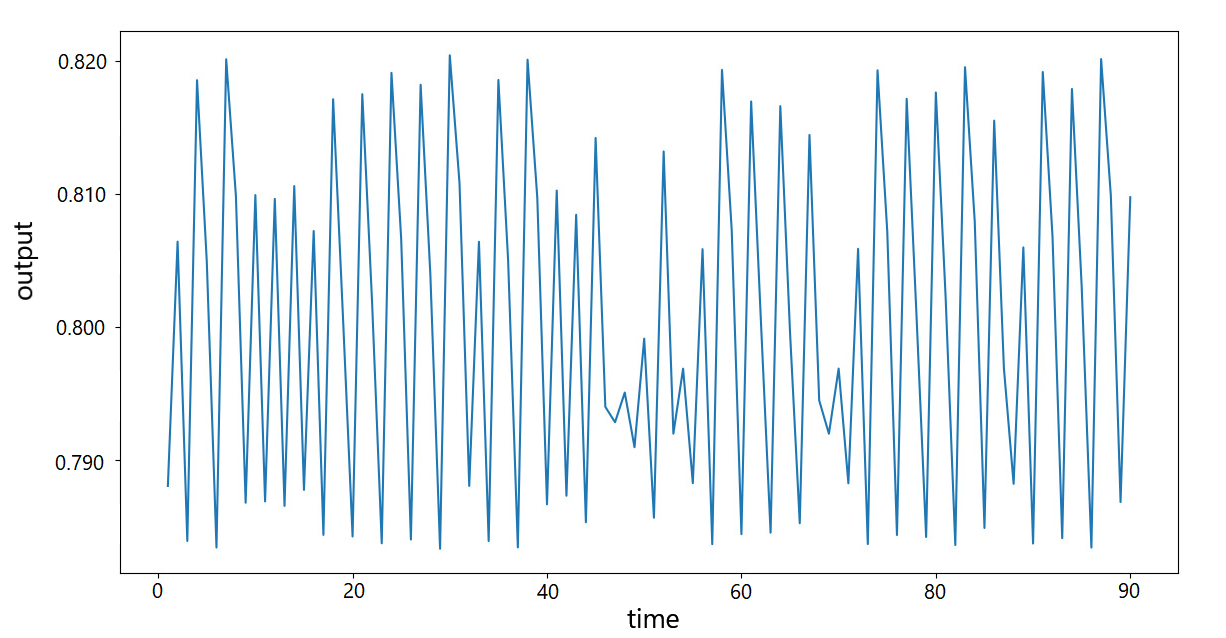}}
	\end{center}
	\caption{Output of the neuron displaying the chaotic dynamical behaviour using flat input data.}\label{fanno}
\end{figure}

\begin{figure}
	\begin{center}
		{\includegraphics[height=4.5cm]{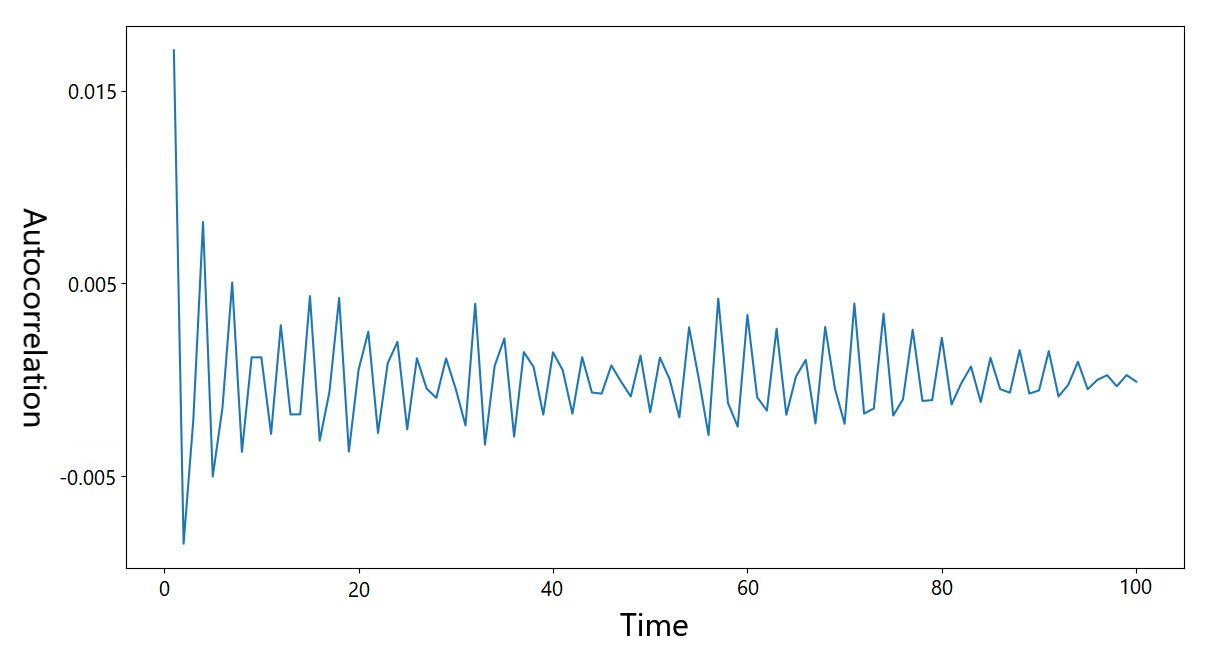}}
	\end{center}
	\caption{Autocorrelation of the resulting output of the neuron.}\label{fauto}
\end{figure}

In the case of the logistic map, when $r = 4$, $\alpha_{min}$ is 0 and $\alpha_{max}$ is 1. These values vary and can be read from the logistic bifurcation diagram as shown in Figure \ref{bifurcation}. The value of $\phi_{max}$ and $\phi_{min}$ depends on the range of the data you want to generate. Increasing the range between the $\phi_{max}$ and $\phi_{min}$ will increase the range of the neuron's output.
For example, with $\alpha_{max} = 1$ and $\alpha_{min}=0$, the function $\phi(t)$ is \begin{equation}\label{enorm}
\phi(t) = \phi_{min} + \alpha(t)(\phi_{max}-\phi_{min})
\end{equation}
Figure \ref{fanno} shows the output from a neuron displaying the chaotic behaviour using $\phi_{max} = 1.1$ and $\phi_{min} = 0.9$. The standard deviation corresponding to this range of $\phi_{max}$ and $\phi_{min}$ is found to be $0.013$. Figure \ref{standard} shows the relationship between the standard deviation of the output generated and the difference between $\phi_{max}$ and $\phi_{min}$. By increasing the range of the bounds $(\phi_{max} - \phi_{min})$, the chaotic behaviour of the resulting output of the neuron varies with larger deviations around the mean, e.g. for $\phi_{max} = 1.2$ and $\phi_{min} = 0.8$, the standard deviation increases proportionately to $0.026$. Figure \ref{fauto} shows the autocorrelation of the resulting output of the neuron, which hovers around zero value, confirming the ability of the modified activation function to display chaotic behaviour.

\begin{figure}
	\begin{center}
		{\includegraphics[height=5.4cm]{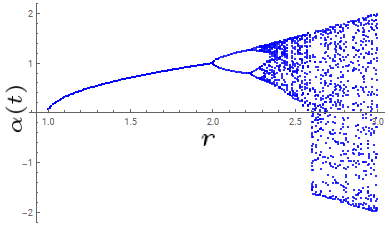}}
	\end{center}
	\caption{Bifurcation diagram of the cubic map.}\label{cubicbifurcation}
\end{figure}

The logistic map is not the only chaotic map that can be used in the proposed activation function. For example, we can use the cubic map, $x_{t+1} = rx_t - x_{t}^3$, which also produces chaotic behaviour, for $2.3 < r < 3$, as shown in Figure \ref{cubicbifurcation}. 

As described earlier, $\alpha_{min}$ and $\alpha_{max}$ need to be set correctly by checking the range of the bifurcation diagram for the corresponding chaotic function being used. It can be worked out by looking at the maximum and minimum value of the graph for certain $r$ values. Table \ref{rtable} gives $\alpha_{min}$ and $\alpha_{max}$ for different values of $r$ for the logistic map, $x_{t+1} = rx_t(1-x_t)$. Table \ref{rtable2} gives the $\alpha_{min}$ and $\alpha_{max}$ values for the cubic map, $x_{t+1} = rx_t - x_t^3$. Both of these are rough estimates measured by generating time series data for different values of $r$, which can be improved by generating longer time series data.
\begin{table}
	\begin{tabular}{|c|c|c|}
		\hline
		$r$   & $\alpha_{min}$ & $\alpha_{max}$ \\
		\hline\hline
		3.5 & 0.382          & 0.875          \\
		\hline
		3.6 & 0.333          & 0.894          \\
		\hline
		3.7 & 0.261          & 0.923          \\
		\hline
		3.8 & 0.181          & 0.949          \\
		\hline
		3.9 & 0.123          & 0.967          \\
		\hline
		4.0 & 0.000          & 1.000          \\
		\hline                 
	\end{tabular}
	\caption{$\alpha_{max}$ and $\alpha_{min}$ for different values of $r$ in the logistic map.}\label{rtable}
\end{table}
\begin{table}
	\begin{tabular}{|c|c|c|}
		\hline
		$r$   & $\alpha_{min}$ & $\alpha_{max}$ \\
		\hline\hline
		2.3 & 0.668          & 1.342          \\
		\hline
		2.4 & 0.585          & 1.408          \\
		\hline
		2.5 & 0.286          & 1.520          \\
		\hline
		2.6 & -1.605         & -0.035         \\
		\hline
		2.7 & -1.698         & 1.664          \\
		\hline
		2.8 & -1.759         & 1.405          \\
		\hline
		2.9 & -1.884         & 1.899          \\
		\hline
		3.0 & -1.953         & 1.861          \\
		\hline
	\end{tabular}
	\caption{$\alpha_{max}$ and $\alpha_{min}$ for different values of $r$ in the cubic map.}\label{rtable2}
\end{table}

\section{Numerical Experiment}\label{ann}
\begin{figure}
	\begin{center}
		{\includegraphics[height=4cm]{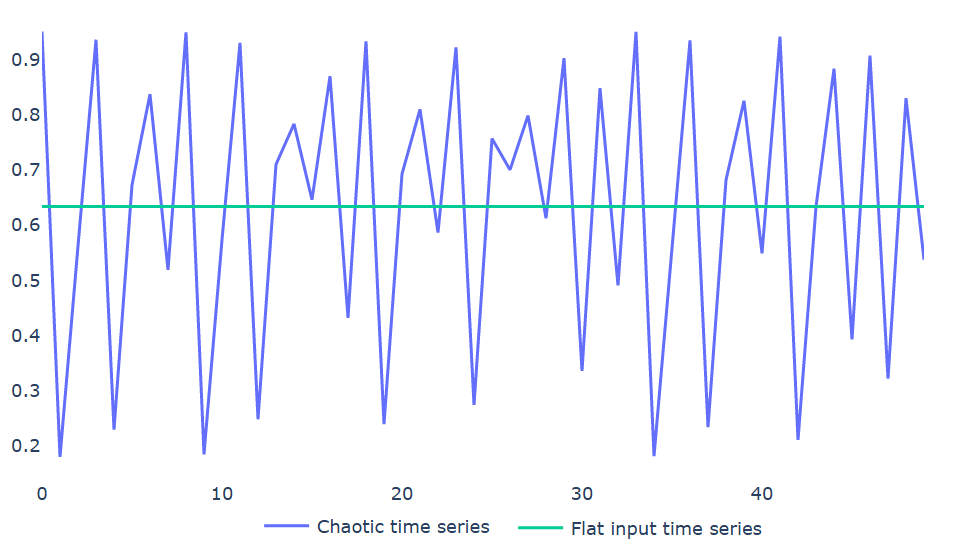}}
	\end{center}
	\caption{Chaotic time series and flat input data.}\label{outputvsinput}
\end{figure}

A numerical experiment was done with a flat time series as the input and chaotic time series as the output, as shown in Figure \ref{outputvsinput}. A single neuron using the proposed spatio-temporal activation function was trained against this chaotic data, while the input was a flat value, with parameters $\phi_{max} = 3.5$ and $\phi_{min} = -2.7$ while $\alpha_{max}$ and $\alpha_{min}$ were 1 and 0 respectively.

\begin{figure}
	\begin{center}
		{\includegraphics[height=5cm]{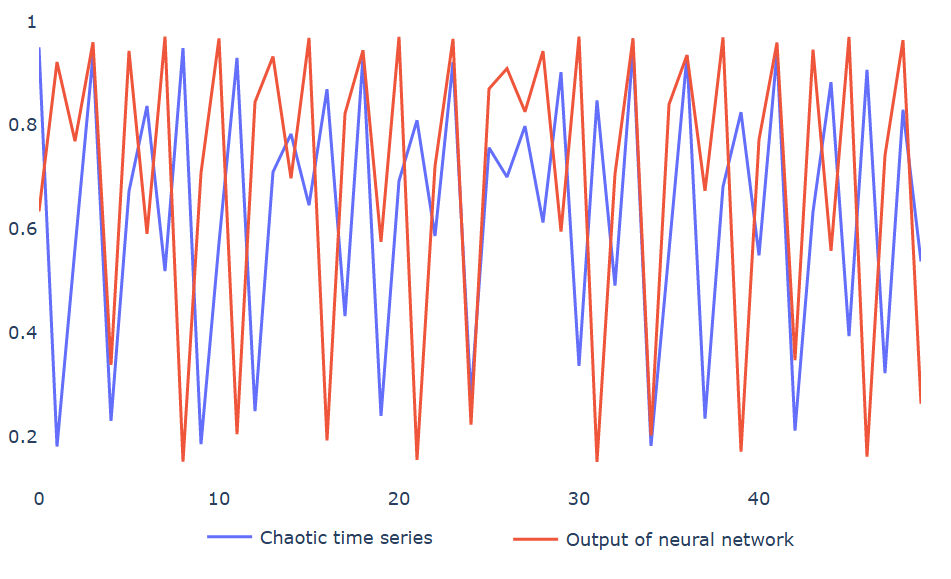}}
	\end{center}
	\caption{Chaotic time series vs Output of neural network.}\label{output}
\end{figure}

After training, the same flat input data was fed to the neural network and the generated output time series data compared well with the actual chaotic time series data as shown in Figure \ref{output}. 

\begin{figure}
	\begin{center}		{\includegraphics[height=4cm]{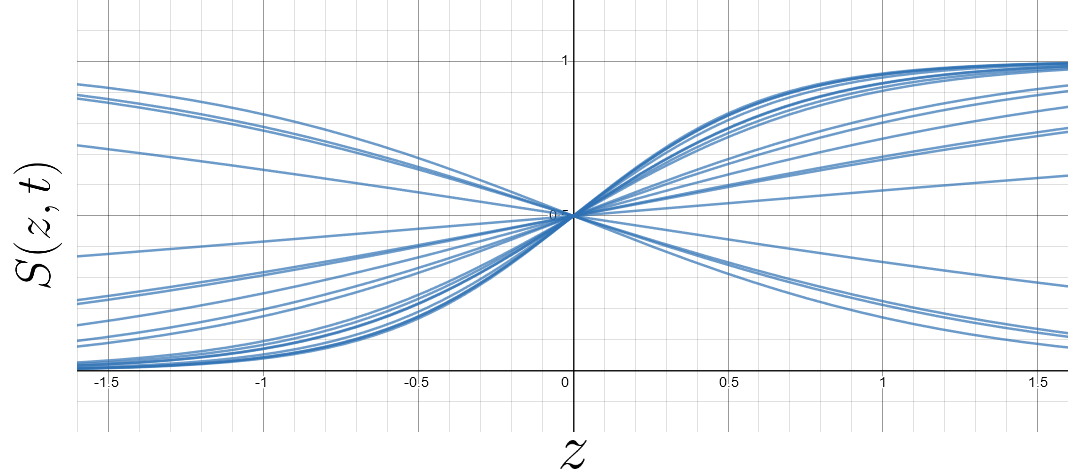}}
	\end{center}
	\caption{Instances of activation function with different $\phi$ values during the generation of the chaotic time series output.}\label{sigmoid}
\end{figure}

In this study, the Lyapunov exponent\cite{salis}, $\lambda$, is used as measure of chaos. The Lyapunov exponent of the function $f(x)$ is defined as follows:
\begin{equation}
\lambda = \frac{1}{n} \sum_{i=0}^{n-1}ln(|f'(x_i)|)
\end{equation}
where $x_{n+1} = f(x_n)$. For the chaotic data, the Lyapunov exponent, $\lambda = -1.01$, while the neural network's output's $\lambda = -0.98$. Thus, the new activation function is able to map the chaotic time series data successfully.\\
Figure \ref{sigmoid} shows the different instances of the activation function while generating the output time series as depicted in Figure \ref{output}. An important thing to notice is that the activation function occasionally flips because the $\phi_{min}$ taken in this example is negative. Though it may be odd, this helps the neuron generate chaotic data that matches the range of the expected output.\\

\section{Conclusion}\label{conclusion}
In the present research work, a two-dimensional activation function is proposed, by including an additional temporal term to impart dynamic behaviour on the output without relying on recurrent neural networks. The temporal term can be either a logistic/cubic map or any other chaotic map. The derivative at any instance of the activation function is similar to the Gaussian probability distribution with a varying, time-dependent standard deviation. Consequently, when the temporal term changes with time, the activation function fluctuates, leading to dynamical behaviour in the output. The new activation function is able to successfully map the chaotic data.

Further research is required to investigate the spatio-temporal activation function in a multilayered feed forward neural network and its ability to capture chaos using logistic/cubic map.


\end{document}